\documentclass[journal]{IEEEtran}

\usepackage{epsfig,graphicx,amsfonts,xcolor,multirow,amsmath,hyperref}

\begin{document}

\title{Richly Activated Graph Convolutional Network for Robust Skeleton-based Action Recognition}

\author{
    Yi-Fan~Song,
    Zhang~Zhang,~\IEEEmembership{Member,~IEEE,}\\
    Caifeng~Shan,~\IEEEmembership{Senior Member,~IEEE,}
    and~Liang~Wang,~\IEEEmembership{Fellow,~IEEE}

    \IEEEcompsocitemizethanks{
        \IEEEcompsocthanksitem This work is sponsored by National Key R\&D Program of China (No.2016YFB1001002), National Natural Science Foundation of China (No.61525306, No.61633021, No.61721004), Shandong Provincial Key Research and Development Program (Major Scientific and Technological Innovation Project) (No.2019JZZY010119) and CAS-AIR.
        \IEEEcompsocthanksitem Yi-Fan Song, Zhang Zhang, and Liang Wang are with the School of Artificial Intelligence, University of Chinese Academy of Sciences (UCAS), Beijing 100190, China, and also with the Center for Research on Intelligent Perception and Computing (CRIPAC), National Laboratory of Pattern Recognition (NLPR), Institute of Automation, Chinese Academy of Sciences (CASIA), Beijing 100190, China. (Email: yifan.song@cripac.ia.ac.cn, zzhang@nlpr.ia.ac.cn, wangliang@nlpr.ia.ac.cn)
        \IEEEcompsocthanksitem Caifeng Shan is with the College of Electrical Engineering and Automation, Shandong University of Science and Technology (SDUST), Qingdao 266590, China, and also with the Artificial Intelligence Research, Chinese Academy of Sciences (CAS-AIR), Beijing 100190, China. (Email: caifeng.shan@gmail.com)
    }
}

\markboth{Journal of IEEE Transactions on Circuits and Systems for Video Technology}
{Song \MakeLowercase{\textit{et al.}}: Richly Activated Graph Convolutional Network for Robust Skeleton-based Action Recognition}

\maketitle

\begin{abstract}
    Current methods for skeleton-based human action recognition usually work with complete skeletons. However, in real scenarios, it is inevitable to capture incomplete or noisy skeletons, which could significantly deteriorate the performance of current methods when some informative joints are occluded or disturbed. To improve the robustness of action recognition models, a multi-stream graph convolutional network (GCN) is proposed to explore sufficient discriminative features spreading over all skeleton joints, so that the distributed redundant representation reduces the sensitivity of the action models to non-standard skeletons. Concretely, the backbone GCN is extended by a series of ordered streams which is responsible for learning discriminative features from the joints less activated by preceding streams. Here, the activation degrees of skeleton joints of each GCN stream are measured by the class activation maps (CAM), and only the information from the unactivated joints will be passed to the next stream, by which rich features over all active joints are obtained. Thus, the proposed method is termed richly activated GCN (RA-GCN). Compared to the state-of-the-art (SOTA) methods, the RA-GCN achieves comparable performance on the standard NTU RGB+D 60 and 120 datasets. More crucially, on the synthetic occlusion and jittering datasets, the performance deterioration due to the occluded and disturbed joints can be significantly alleviated by utilizing the proposed RA-GCN.\footnote{The codes and pretrained models of the preposed RA-GCN are available at \href{http://github.com/yfsong0709/RA-GCNv2}{http://github.com/yfsong0709/RA-GCNv2}.}
\end{abstract}

\begin{IEEEkeywords}
    Action Recognition, Skeleton, Activation Map, Graph Convolutional Network, Occlusion, Jittering
\end{IEEEkeywords}

\section{Introduction}
\label{sec:intro}

Human action recognition has achieved promising progress in recent computer vision researches and plays an increasingly crucial role in many potential applications, such as video surveillance, human-computer interaction, video retrieval and so on \cite{aggarwal2011human, sudha2017approaches, weinland2011survey}. The main purpose of action recognition is to classify human actions from motion data which can be captured as RGB videos \cite{sharma2015action, tran2015learning}, depth maps \cite{jalal2017robust}, infrared images \cite{akula2018deep} and 3D skeleton sequences \cite{liu2016spatio, zhang2017view, yan2018spatial}.

Traditional action recognition is dominated by RGB video-based methods. These methods usually consider RGB videos as temporal sequences of image frames, and use sequential models such as recurrent neural network (RNN) to exploit temporal information from all the feature maps extracted by convolution neural networks (CNN) for each frame \cite{sharma2015action}. On the other hand, many researchers utilize 3D CNN to derive useful information directly from the videos \cite{tran2015learning}, which has obtained a comparable performance with the former methods. Nevertheless, both of these two categories extract spatial structure information from 2D RGB frames, while the spatial configurations of actors are absolutely presented in 3D space. Thus, these RGB-based methods will lose some crucial information due to the intrinsic weakness. Moreover, the RGB videos often contain complex background and illumination variations, which leads to significant performance degradation in practice.

Compared to RGB videos, skeleton-based human action recognition methods reflect a growing prospect, due to its superiority in background adaptability, robustness to light variations and less computational cost. Skeleton data is composed of 2D/3D coordinates of multiple skeleton joints in motion sequences, which can be either collected by multimodal sensors such as Kinect or directly estimated from 2D images by pose estimation methods \cite{cao2017realtime}. Current methods usually deal with skeleton data in two ways. One is to connect these joints into a global vector, then model temporal information by using RNN-based methods \cite{liu2016spatio, zhang2017view, lee2017ensemble, si2018skeleton}. The other way is to treat or expand temporal sequences of joints into 2D images, then utilize CNN-based methods to recognize actions \cite{li2017skeleton, soo2017interpretable, li2018co, tang2018deep}. However, it is difficult to utilize the spatial structure information among skeleton joints effectively with both the RNN and CNN methods, though many researchers propose additional constraints or dedicated network structures to model the spatial structure of skeleton joints. Recently, graph neural networks (GNN), which can explicitly incorporate graphical structure information into the learning of neural network, have made great progress in many fields \cite{kipf2016semi}. Yan et al.\cite{yan2018spatial} firstly propose a spatial temporal graph convolutional network (ST-GCN) to capture the patterns embedded in the spatial configuration as well as the temporal dynamics in skeleton sequences, which achieves a significant improvement in action recognition.

\begin{figure}[t]
    \centerline{\includegraphics[width=8.5cm]{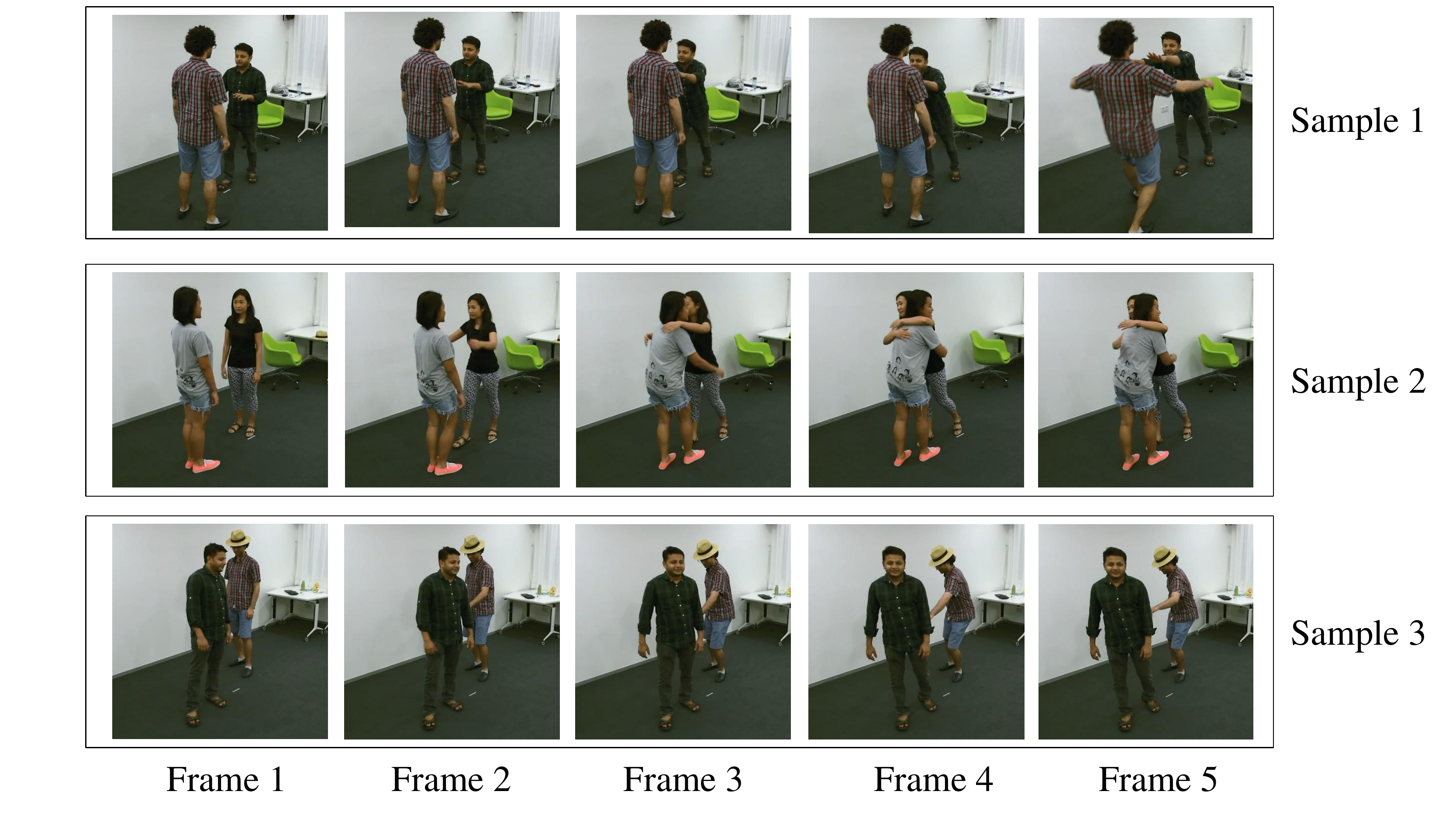}}
    \caption{Some examples of occluded actions. The person behind is partially occluded by the person in front.}
    \label{fig:occlusion}
\end{figure}

However, current skeleton-based action recognition models still lack of robustness to the noisy or incomplete skeleton data captured in real scenarios. For example, the subjects may be self-occluded by pose variations or occluded by other contextual objects. Fig.\ref{fig:occlusion} displays some examples of occluded actions due to other persons in the scenes. The noisy data will deteriorate the performance of the models heavily. Therefore, how to enhance the robustness of skeleton-based action recognition models is still an urgent and challenging problem.

To against various degradations, ensemble learning has proved to be an effective strategy \cite{krawczyk2017ensemble}, which induces multiple classifiers based on the same or distinct predictive classifiers so that the integration of these individual classifiers could enhance the robustness of recognition models. Inspired by the success of ensemble learning, in this paper, we propose a GCN-based multi-stream model, which aims to learn rich discriminative features from skeleton motion sequences, and thereby improve the robustness of the proposed model. The purpose of each stream in our approach is to explore a group of discriminative features over the skeleton joints unactivated by previous streams. The learnt redundant but complementary features over all skeleton joints provide an effective strategy to handle the noisy or incomplete skeleton data. For example, when we recognize the action {\it throwing}, the most discriminative joints are located on the two arms at the moment of object leaving the hands in the process of {\it throwing}, while the body swaying as well as the contextual sub-actions of hands can also be used to infer the action of {\it throwing}.

In order to distinguish the most informative joints for each stream, we introduce a successful technique named class activation maps (CAM) \cite{zhou2016learning} into our model, which initially aims to visualize the activation heatmap in the final CNN layer responsible for visual classification. The activation maps obtained by previous GCN streams are accumulated as a mask matrix to inform the new stream about which joints have been already activated. Then, the new stream will be forced to explore new discriminative features from unactivated joints. Therefore, the proposed method is called richly activated GCN (RA-GCN), where the richly discovered and complementary features will improve the robustness of the model to non-standard skeletons. To the best of our knowledge, this is the first time to employ the CAM technique to enhance the model robustness by expanding the activated skeleton joints, which alleviates the problems of occlusion and jittering in skeleton-based action recognition.

To validate the advantages of the proposed methods, besides the traditional skeleton action datasets, the NTU RGB+D 60 \cite{shahroudy2016ntu} \& 120 \cite{liu2019ntu} datasets, we also build four synthetic occlusion datasets, where the joints in the NTU 60 and 120 datasets are partially occluded over both spatial and temporal dimensions, and two synthetic jittering datasets, where some randomly selected joints are disturbed by Gaussian noises. More details of these datasets can be found in Section \ref{ssec:dataset}. Our experiments on these new datasets demonstrate that the proposed RA-GCN significantly alleviates the performance deterioration in the case of incomplete or noisy skeleton data.

This work is an extension of an earlier and preliminary version presented in \cite{song2019richly}. Compared to our previous work, the modifications and contributions of this paper are summarized as follows:
\begin{itemize}
    \item In previous work, the activation masks are obtained by a Softmax function in the activation modules, which activates only a few joints for each stream. In contrast, we propose to use a normalization activation function to expand the activated scope, thus the corresponding stream will obtain a better and more interpretable activation map.
    \item Compared to previous work, we extend the original loss function with a number of additional cross-entropy regularizations on each individual network stream, so that the features can be learnt more effectively.
    \item The synthetic datasets are extended by more degradation operators, where the occlusion degradation is further divided into four types, including Frame, Part, Block and Random, and two synthetic jittering datasets are newly constructed. More experiments are performed to validate the effectiveness and robustness of the proposed approach in different degradation conditions.
\end{itemize}

The remainder of this paper is organized as follows: Section \ref{sec:related} describes recent studies related to our work. Section \ref{sec:model} introduces several crucial components of the proposed RA-GCN. Extensive experimental results on standard and non-standard datasets are reproted in Section \ref{sec:experiments}, and the conclusion of this paper is given in Section \ref{sec:conclusion}.

\section{Related Work}
\label{sec:related}

\begin{figure*}[t]
    \centering
    \includegraphics[width=18cm]{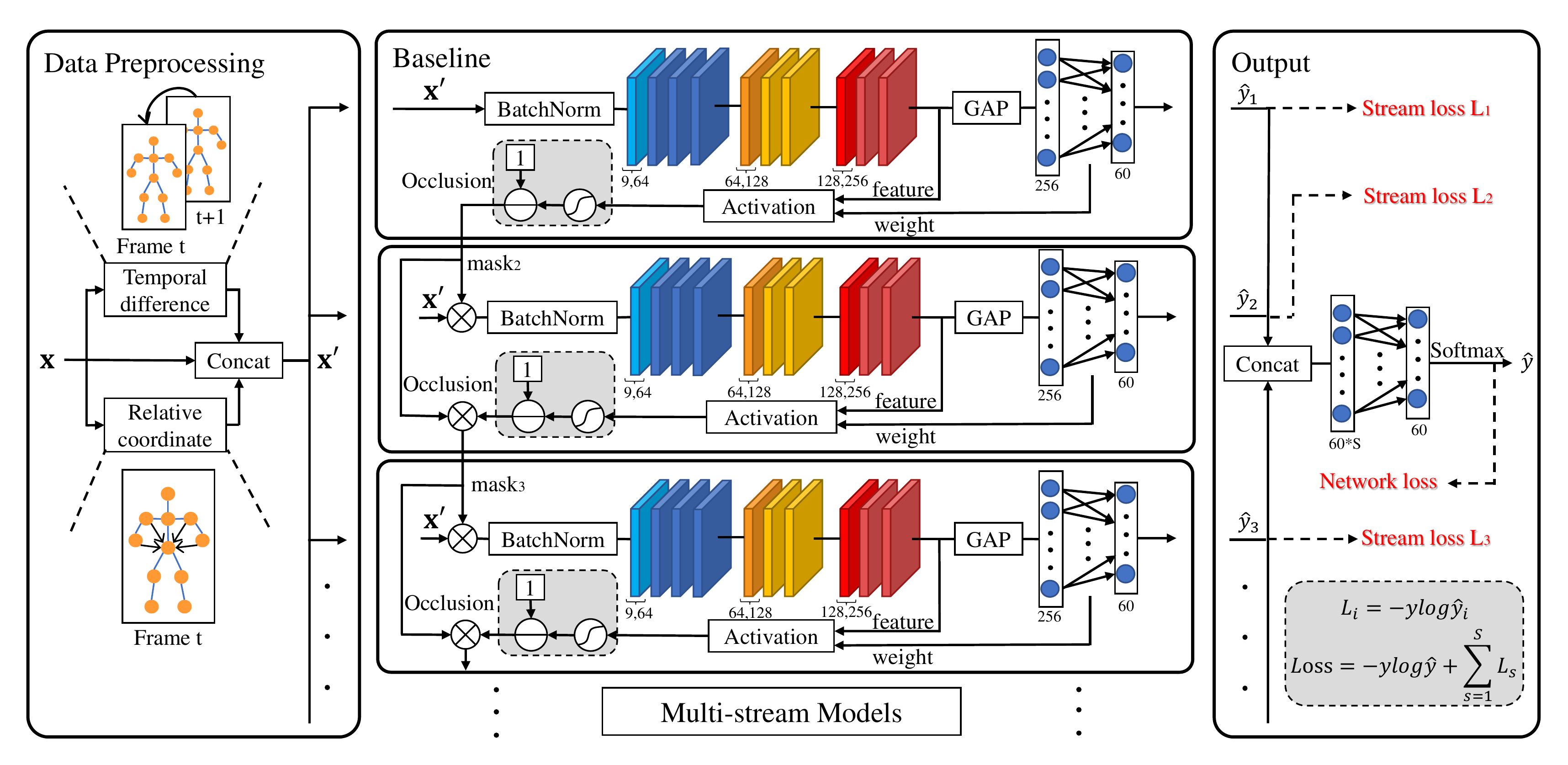}
    \caption{The pipeline of RA-GCN with three stream networks. Each stream contains one ST-GCN. The two numbers under the ST-GCN layers are the numbers of input and output channels, respectively. Other layers contain the same input and output channels. Both of the fifth and the eighth layers use a temporal stride 2 to reduce the sequence length. {\bf GAP} and {\bf Concat} are global average pooling and concatenation operation, $S$ is the number of streams, and $\otimes$ and $\ominus$ denote element-wise multiplication and subtraction, respectively. This model totally contains three steps. The input sequence $\bf x$ is firstly transformed into $\bf x'$ by the data preprocessing module. Secondly, $\bf x'$ will be sent to each stream after being filtered by a corresponding mask matrix. Finally, the output of each stream will be concatenated to obtain the final class of $\bf x$.}
    \label{fig:pipeline}
\end{figure*}

\paragraph{Skeleton-based models} To find a more effective representation of the dynamics of human actions, Johansson \cite{johansson1973visual} utilizes 3D skeleton sequences for action recognition, making an obvious decrease of computational cost as well as a good performance boost. Recently, with the rapid development of deep learning techniques, skeleton-based action recognition methods have attracted increasing attentions. Researchers have proposed various models to improve the performance of action recognition, which can be divided into three major categories. The first category builds the models with convolutional networks. For example, Li et al. \cite{li2018co} propose a CNN-based co-occurrence feature learning framework, which gradually aggregates various levels of contextual information. Kim et al. \cite{soo2017interpretable} build a temporal convolutional network to explicitly learn readily interpretable spatio-temporal representations for 3D human action recognition.

Besides, for the second category, researchers concatenate all joints in one frame into a single vector, then use sequential models such as long short-term memory (LSTM) to explore the temporal dynamics. Du et al. \cite{du2015hierarchical} design a hierarchical bidirectional RNN to capture rich dependencies between different human body parts. The study in \cite{zhang2017view} employs a view adaptive LSTM, which enables the network itself adaptive to the most suitable observation viewpoints. Additionally, Song et al. \cite{song2017end} firstly introduce attention modules into skeleton-based action recognition.

Both CNN-based and RNN-based methods are still limited to extract the spatial structure information among skeleton joints, where the joints of different body parts are connected as a skeleton graph. Instead, in the third category, graph-based methods can be naturally utilized to deal with the skeleton graph, which successfully captures the most informative features for various actions. Si et al. \cite{si2018skeleton} use GNN to model the relationships among five body parts. Yan et al. \cite{yan2018spatial} initially introduce GCN into skeleton-based action recognition, and produce a baseline named ST-GCN for future research. Based on the ST-GCN, many studies achieve continuous improvements on skeleton-based action recognition \cite{si2019attention, thakkar2018part, shi2019two}.

\paragraph{Occlusion in human action recognition} Occlusion is a prominent challenge in human action recognition. If the skeleton joints are partially occluded, the approaches mentioned above will face a considerable decline of performance. To handle this problem, Wang et al. \cite{wang2009hog} try to infer occlusion maps from a global SVM classifier, and Weinland et al. \cite{weinland2010making} propose a local partitioning and hierarchical classification of the 3D Histogram of Oriented Gradients (HOG) descriptor for providing the robustness to both occlusions and view point changes. However, there are few studies addressing the problem of noisy or incomplete data in skeleton joints. In this paper, we propose an approach to exploring rich features over all joints, so as to alleviate the effects of data degradation.

\paragraph{Salient Regions Exploration} Similar with our motivation, some previous studies have proposed to explore the salient regions or erase them to exploit complementary information for referring expression grounding \cite{hou2018self, liu2019improving} or weakly supervised detection tasks \cite{li2018tell}. The study \cite{hou2018self} proposes a simple yet effective network to prohibit attentions from spreading to unexpected background regions, in order to promote the quality of object attention. Liu et al. \cite{liu2019improving} design a novel attention-guided erasing approach to aligning various types of information crossing visual and textual modalities. Moreover, Li et al. \cite{li2018tell} provide a framework to dynamically erase the focused area according to the on-line attention maps. However, previous salient regions exploration methods mainly concentrate on object detection or localization tasks in images, while in this work, we exploit the complementary attended skeleton joints for alleviating the occlusion or jittering problems, which is still not considered in previous work.

\section{Model Architecture}
\label{sec:model}

In order to enhance the robustness of action recognition models, we propose the RA-GCN to explore sufficient discriminative features from all skeleton joints. The proposed RA-GCN constructs a multi-stream network, where each stream is responsible for extracting features from a group of activated joints. In this way, when the joints activated by the first stream are occluded, the model can also discover discriminative information from the other streams. The overview of RA-GCN is presented in Fig.\ref{fig:pipeline}. Suppose that $V$ is the number of joints in one skeleton and $T$ is the number of frames in one sequence, the size of input data $\bf x$ is $C_{in}\times T\times V$, where $C_{in}=3$ denotes the 3D coordinates of each joint. Note that different skeletons in a multi-agent action are treated as different samples.

The proposed method consists of three main steps. Firstly, in the preprocessing module, for extracting more informative features, the input data $\bf x$ is transformed into $\bf x'$, which is subsequently sent to all the GCN streams. Secondly, for each stream, the skeleton joints in $\bf x'$ will be filtered by the element-wise product with a mask matrix, which records the currently unactivated joints. These joints are distinguished by accumulating the activated maps calculated by the activation modules of preceding streams. Here, the mask matrix of each stream is initialized to an all-one matrix with the same shape as $\bf x'$. After the masking operation, the input data of each stream only contains the joints unactivated by the preceding streams, and passes through a baseline network to obtain a feature representation based on the incomplete skeleton joints. Finally, the features of all streams are concatenated in the output module, and a fully connected layer with Softmax activation function is used to obtain the final class of input $\bf x$. These three steps will be discussed in details in next sections.

\subsection{Data Preprocessing}
\label{ssec:preprocessing}

\begin{figure}[t]
    \centering
    \includegraphics[width=8.5cm]{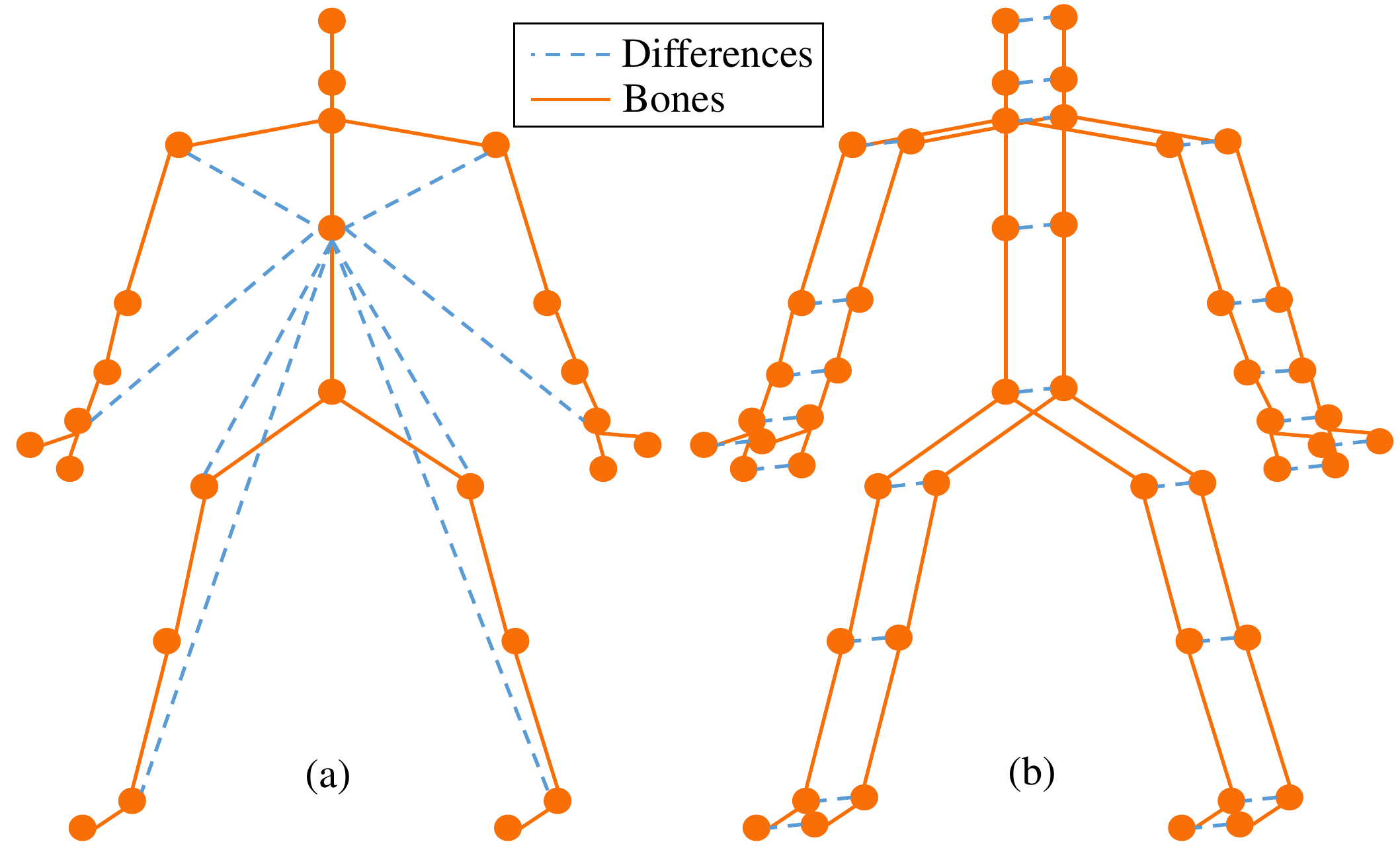}
    \caption{(Best viewed in color.) The illustration of data preprocessing. (a) is relative joint coordinates. (b) is temporal displacements. The {\color{orange}orange} solid lines mean the physical connections between neighboring joints (bones) and the {\color{blue}blue} dash lines denote the differences between two joints.}
    \label{fig:preprocessing}
\end{figure}

Usually, some actions such as {\it taking off the shoes} and {\it wearing the shoes} are extremely similar with only spatial features. To tackle this problem, conventional RGB-based methods introduce a sophisticated technique named Optical Flow \cite{wang2016temporal} into their models, for depicting the motion features exactly. Besides, the work in \cite{thakkar2018part} argues that relative coordinates of joints are usually more informative than absolute coordinates. Inspired by this, geometric features such as relative coordinates and motion features such as temporal displacements are applied in our models to increase the discriminative information for action recognition. Therefore, the input data need to be preprocessed before distributing it to all the GCN streams.

The relative coordinates can be recognized as the difference ${\bf \dot{x}}_r$ between all joints and the center joint (middle spine) in each frame, which can be seen in Fig.\ref{fig:preprocessing}(a). In this way, all joints are transformed to the relative coordinates, which is more robust to the changing position. Besides, for extracting more informative motion features, we compute ${\bf \dot{x}}_t$ by ${\bf x}_{t+1}-{\bf x}_{t}$, where ${\bf x}_t$ means the feature map of the $t^{th}$ frame, which is shown in Fig.\ref{fig:preprocessing}(b). Then, ${\bf x'}$ will be obtained by concatenating ${\bf x}$, ${\bf \dot{x}}_r$ and ${\bf \dot{x}}_t$.

\subsection{Richly Activated GCN}
\label{ssec:methods}

\subsubsection{Baseline Model}
\label{sssec:baseline}

The baseline of our method is the ST-GCN \cite{yan2018spatial}, which is composed of ten graph convolutional layers. Yan et al. \cite{yan2018spatial} formulate spatial graph convolutional operation as follows:
\begin{equation}
    \label{eq:spatial}
    f_{out}(v_{ti}) = \sum_{v_{tj}\in B(v_{ti})}\frac{1}{Z_{ti}(v_{tj})} f_{in}(v_{tj})\cdot {\bf w}(l_{ti}(v_{tj})),
\end{equation}
where $f_{in}$ and $f_{out}$ are the input and output feature maps respectively, $v_{ti}$ denotes the $i^{th}$ joint at the $t^{th}$ frame, which can also be regarded as the root joint in this procedure, $B(v_{ti})$ is the neighbor set of $v_{ti}$, the normalizing term $Z_{ti}$ is added to balance the contributions of different neighbors, ${\bf w}(\cdot)$ is a weight function implemented by several 1$\times$1 Conv layers and $l_{ti}(\cdot)$ means a label function. There are three label functions in \cite{yan2018spatial}, but we only choose the distance-based label function in our method, which defines $l_{ti}(v_{tj}) = d(v_{ti},v_{tj})$. That means the neighbor set $B(v_{ti})$ is divided into several subsets, according to the graph distance between $v_{tj}$ and the root joint $v_{ti}$. For example, if the joint $v_{tj}$ directly connects with the root joint $v_{ti}$, then $d(v_{ti},v_{tj})=1$. The joints with the same distance will form a subset and share a learnable weight function ${\bf w}(\cdot)$. To implement the spatial graph convolution with the adjacency matrix $\bf A$, Eq.\ref{eq:spatial} is transformed into:
\begin{equation}
    {\bf f}_{out} = \sum_{d=0}^{D} {\bf W}_d{\bf f}_{in} ({\bf \Lambda}_d^{-\frac{1}{2}}{\bf A}_d{\bf \Lambda}_d^{-\frac{1}{2}}\otimes {\bf M}_d),
\end{equation}
where $D$ is the predefined maximum distance, ${\bf A}_d$ denotes the adjacency matrix for distance $d$, $\Lambda_d^{ii} = \sum_k A_d^{ik} + \alpha$ is the normalized diagonal matrix, $A_d^{ik}$ denotes the element of the $i^{th}$ row and $k^{th}$ column of ${\bf A}_d$ and $\alpha$ is set to a small value, e.g., $10^{-4}$, to avoid the empty rows in ${\bf \Lambda}_d$. For each adjacency matrix, we accompany it with a learnable matrix ${\bf M}_d$, which expresses the importance of all edges in one skeleton.

\begin{figure}[t]
    \centering
    \includegraphics[width=8.5cm]{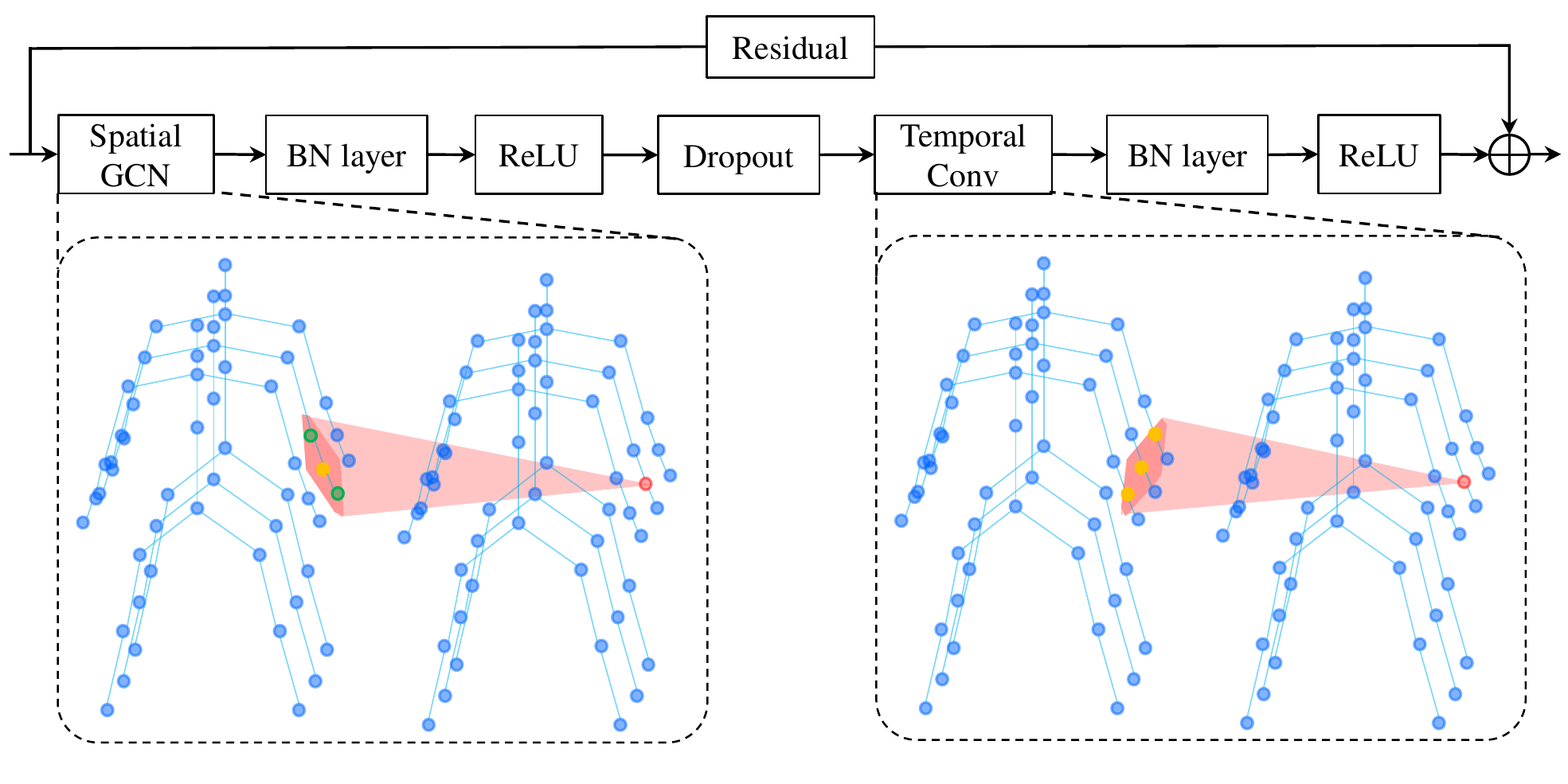}
    \caption{The structure of an ST-GCN layer, where BN means the BatchNorm layer, and $\oplus$ denotes the element-wise summation operation.}
    \label{fig:stgcn}
\end{figure}

After the spatial graph convolutional block, a $1\times L$ convolutional layer is used to extract temporal information of the feature map ${\bf f}_{out}$, where $L$ is the temporal window size. Both spatial and temporal convolutional blocks are followed with a BatchNorm layer and a ReLU layer, and the total ST-GCN layer contains a residual connection. Besides, a dropout layer with the drop probability of 0.5 is added between every spatial convolutional block and temporal convolutional block to avoid overfitting. The structure of one ST-GCN layer is shown in Fig.\ref{fig:stgcn}.

\subsubsection{Activation Module}
\label{sssec:activation}

The activation module in the RA-GCN is constructed to distinguish the activated joints of each stream, then guide the learning process of the new stream by accumulating the activated maps of preceding streams. This procedure can be mainly implemented by extending the CAM technique \cite{zhou2016learning} to the field of GCN. The original CAM technique is to localize class-specific image regions in CNNs, and $score_c$ is defined as the scores of all pixels for class $c$, where the score of each pixel is
\begin{equation}
    score_c(x,y)=\sum_k w_k^c f_k(x,y).
\end{equation}
In this formulation, $f_k(\cdot,\cdot)$ is the feature map before the global average pooling operation, and $w_k^c$ is the weight of the $k^{th}$ channel for class $c$. In this paper, we replace the coordinate $(x,y)$ in a feature map with the frame number $t$ and the joint number $i$ in a skeleton sequence, by which we are able to locate the activated joints. Here, the class $c$ is selected as the ground truth. We use $score_c^s$ to denote the score map of all joints for the true class and the $s^{th}$ stream. To determine which joints are activated by the corresponding stream, a predefined threshold $\delta$ is utilized, and the activation map of $map_c^s$ is calculated by
\begin{equation}
    map_c^s=\varepsilon(\frac{score_c^s}{max(score_c^s)}-\delta)
\end{equation}
where $\varepsilon(\cdot)$ is the Heaviside step function and $max(\cdot)$ denotes the maximum function. Then, the mask matrix of the $s^{th}$ stream is represented as
\begin{equation}
    \label{eq:unactivated}
    mask_s=(\prod_{i=1}^{s-1}mask_i)\otimes(1-map_c^{s-1}),
\end{equation}
where $\prod$ denotes the element-wise product of all mask matrices before the $s^{th}$ stream. Specially, the mask matrix of the first stream is an all-one matrix. Finally, the input of the $s^{th}$ stream will be obtained by
\begin{equation}
    \label{eq:input}
    {\bf x}_s={\bf x'}\otimes mask_s,
\end{equation}
where $\bf x'$ is the skeleton representation after preprocessing.

Eq.\ref{eq:unactivated} and Eq.\ref{eq:input} illustrate that the input of the $s^{th}$ stream only consists of the joints which are not activated by previous streams. Thus, the RA-GCN will explore discriminative features from all joints sufficiently.

\subsubsection{Loss Function}
\label{sssec:loss}

In our previous model \cite{song2019richly}, the loss function only supervises the total network, which is extended with a number of additional losses for each individual network streams in this paper, so that the feature can be learnt more effectively. Suppose that ${\bf \hat{y}}_s\in\mathbb{R}^C$ is the output of the $s^{th}$ stream, ${\bf \hat{y}}\in\mathbb{R}^C$ is the output of the whole model and ${\bf y}\in\mathbb{R}^C$ is the ground truth, where $C$ is the number of classes. Then the loss function of the proposed RA-GCN is
\begin{equation}
\label{eq:loss}
L=-{\bf y}log{{\bf \hat{y}}}-\sum_{s=1}^{S}{\bf y}log{{\bf \hat{y}}_s}
\end{equation}
where $S$ is the number of streams.

\section{Experimental Results}
\label{sec:experiments}

\begin{figure}[t]
    \centering
    \includegraphics[width=8.5cm]{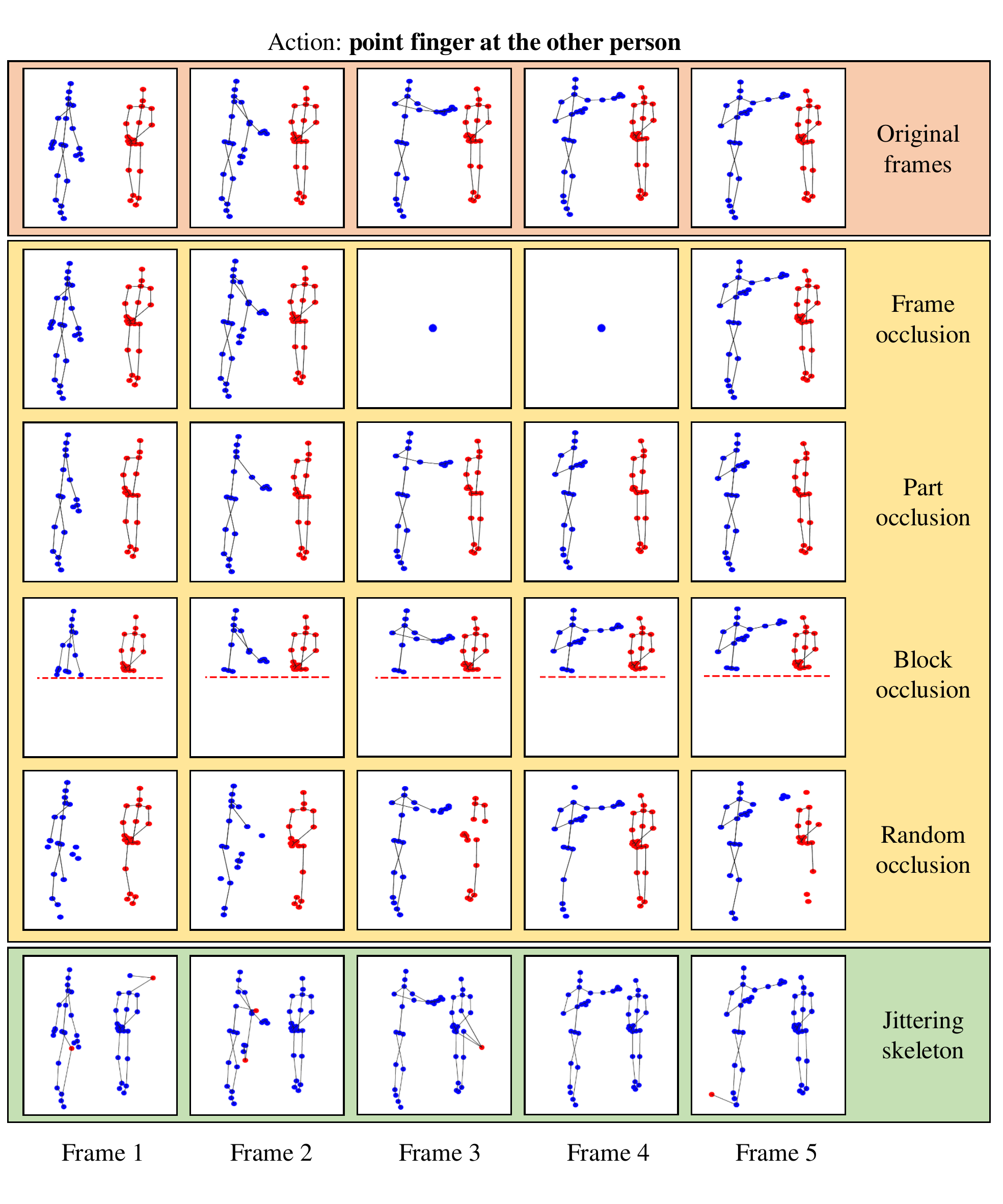}
    \caption{(Best viewed in color.) The demonstration of the degraded datasets generated by adding various types of noise and occlusion to the standard NTU 60 dataset. The first line is the original frames. The middle lines are the frame occlusion without frames 3 and 4, the part occlusion without left arm, the block occlusion without the joints under the red dash line and the random occlusion with an occluded probability $p=0.2$, respectively. In addition, the bottom line is an example of the jittering skeletons, where the {\color{red}red} joints denote the noisy joints.}
    \label{fig:dataset}
\end{figure}

\subsection{Dataset}
\label{ssec:dataset}

\paragraph{NTU RGB+D 60 \cite{shahroudy2016ntu}} This dataset is a large-scale indoor action recognition dataset, which contains 56880 video samples collected by Microsoft Kinect v2, and consists of 60 action classes performed by 40 subjects. Each video is composed of 25 joints and no more than two skeletons in one frame. The maximum frame number $T$ is set to 300 for simplicity. The authors of this dataset recommend two benchmarks:  (1) {\bf cross-subject (CS)} contains 40320 samples for training and 16560 samples for evaluation, by splitting 40 subjects into two groups; (2) {\bf cross-view (CV)} uses cameras 2 and 3 (37920 samples) for training and camera 1 (18960 samples) for evaluation. We follow this convention and report the top-1 recognition rate on both two benchmarks. In addition, according to \cite{lee2017ensemble}, there are 302 wrong samples that need to be ignored during training and evaluation.

\paragraph{NTU RGB+D 120 \cite{liu2019ntu}} This dataset is currently the largest indoor action recognition dataset, which is an extended version of NTU 60 dataset. It contains 114480 videos and consists of 120 classes. Similarly, two benchmarks are suggested: (1) {\bf cross-subject (CSub)} contains 630226 samples for training and 50922 samples for evaluation; (2){\bf cross-setup (CSet)} contains 54471 videos for training and 59477 videos for evaluation, which are separated based on the distance and height of their collectors. Note that there are 532 bad samples in this dataset which should be ignored in all experiments.

\paragraph{Occlusion dataset} To validate the robustness of our method to incomplete skeletons, we construct a synthetic occlusion dataset based on the CS benchmark of NTU 60 dataset and the CSet benchmark of NTU 120 dataset, where some joints are selected to be occluded (set to zero) over both spatial and temporal dimensions. Note that this operation is executed before data preprocessing, and all the joints related to the occluded joints (with zero energy) are ignored in data preprocessing phase. For example, if the frame ${\bf x}_t$ is occluded, then the temporal displacements ${\bf \dot{x}}_t={\bf x}_{t+1}-{\bf x}_t$ and ${\bf \dot{x}}_{t-1}={\bf x}_t-{\bf x}_{t-1}$ are both set to zero. This synthetic dataset consists of four cases, which are frame occlusions, part occlusions, block occlusions and random occlusions, respectively. Part occlusion and block occlusion are both used to simulate the real scenarios that occluded by contextual objects. Frame occlusion is designed for the loss of key frames, while random occlusion is for data missing in signal transmission. Some examples on the four types of occlusions are illustrated in Fig.\ref{fig:dataset}. Note that all the models in occlusion experiments are trained with standard skeletons, and then tested with incomplete skeletons.

\paragraph{Jittering dataset} Skeleton jittering is a common factor that has a big impact on the recognition performance. To claim the robustness of the proposed method to jittering skeletons, we propose a synthetic jittering dataset based on the CS benchmark of NTU 60 dataset and the CSet benchmark of NTU 120 dataset, where the Gaussian noise $N(\mu,\sigma^2)$ is added to some randomly selected joints to simulate the jittering joints. This jittering operation is also executed before data preprocessing module, while the data preprocessing module has no difference with that in standard setting. In this paper, two types of Gaussian noise are used, which are $N(0,0.1^2)$ and $N(0,0.05^2)$. In the bottom line of Fig.\ref{fig:dataset}, an example of jittering skeletons is displayed, and the red joints denote the noisy joints. Similar with the occlusion dataset, all the models of this dataset are trained with standard skeletons.

\subsection{Implementation Details}
\label{ssec:details}

In our experiments, some hyper-parameters need to be modified. The initial learning rate is set to 0.1 and divided by 10 every 20 epochs, while the maximum number of iterations is set to 60. The models are learnt by using the stochastic gradient descent (SGD) algorithm with a momentum 0.9 and a weight decay $10^{-4}$. In order to avoid overfitting, the probability of the dropout layer between the spatial and temporal blocks is selected as 0.5. The first four ST-GCN layers have 64 channels for output, while the number will be 128 and 256 for the middle three layers and the last three layers. Moreover, at the fifth and eighth layers, the temporal convolutional blocks contain a temporal stride 2, for reducing the computational cost. As to the maximum graph distance $D$, the temporal window size $L$ and the mask threshold $\delta$, we will discuss about their effects in Section \ref{ssec:parameters}.

Before training a multi-stream RA-GCN, we need to pretrain a one-stream RA-GCN with preprocessed skeleton data to get the baseline model, so as to ensure that the first stream of RA-GCN is able to capture the most informative joints. Accordingly, the following streams are forced to seek for other discriminative joints. Additionally, the mask matrix of each stream is initialized to an all-one matrix. Finally, we finetune the RA-GCN model with the setting mentioned above. All the experiments are performed on two TITAN X GPUs.

\subsection{Parameters Setting}
\label{ssec:parameters}

\begin{table}[t]
\caption{Comparison of different parameter settings of RA-GCN on the two benchmarks NTU 60 (\%)}
\label{tab:parameters}
\centering
\begin{tabular}{cccccc}
\hline
Model & Parameters & & CS &  & CV \\
\hline
&$D=1, L=5$&  & 85.2 & & 90.5 \\
&$D=2, L=5$&  & {\bf 85.8} & & 91.6 \\
baseline &$D=3, L=5$&  & {\bf 85.8} & & 92.2 \\
(1s RA-GCN*) &$D=1, L=9$&  & 85.4 & & 91.7 \\
&$D=2, L=9$&  & 85.4 & & 92.7 \\
&$D=3, L=9$&  & 85.0 & & {\bf 93.1} \\
\hline
&$\delta=0.1$ $(D=2, L=5)$&  & 86.5 & & -- \\
2s RA-GCN*&$\delta=0.3$ $(D=2, L=5)$&  & {\bf 86.7} & & -- \\
&$\delta=0.5$ $(D=2, L=5)$&  & 86.3 & & -- \\
\hline
\multicolumn{4}{l}{*: 1s and 2s denote the number of streams}
\end{tabular}
\vspace{-0.4cm}
\end{table}

\begin{table}[t]
\caption{Comparison of different model settings on the CS benchmark of NTU 60 (\%)}
\label{tab:ablation}
\centering
\begin{tabular}{cccccc}
\hline
Model & Setting & & accuracy \\
\hline
& w/o activation module &  & 85.5 \\
2s RA-GCN & w/o pretrained &  & 85.2 \\
& w/ activation function in \cite{song2019richly} &  & 85.8 \\
\hline
& only raw skeleton &  & 76.8 \\
2s RA-GCN & only relative coordinates &  & 73.4 \\
& only temporal displacements &  & 83.1 \\
\hline
1s RA-GCN & -- &  & 85.8 \\
2s RA-GCN & -- &  & 86.7 \\
3s RA-GCN & -- &  & {\bf 87.3} \\
4s RA-GCN & -- &  & 87.2 \\
\hline
\end{tabular}
\vspace{-0.4cm}
\end{table}

In order to train a baseline model, we firstly need to determine the value of two hyper-parameters introduced in Section \ref{sssec:baseline}, i.e., the $D$ for the maximum distance and the $L$ for the temporal window size. These two hyper-parameters have a great impact on our model, because they control the receptive field of the GCN blocks. To find the optimal values, we evaluate many groups of the two hyper-parameters ($D\in\{1,2,3\}$ and $L\in\{3,5,7,9,11\}$) on the NTU 60 dataset, and some representative experimental results are given in the first part of Tab.\ref{tab:parameters}. It is observed that the baseline model achieves the best accuracy when $D=2$ and $L=5$ on the CS benchmark. As to the CV benchmark, $D$ and $L$ are optimally set to 3 and 9, respectively. Note that it is not always better to choose bigger $D$ and $L$, since a bigger receptive field will lead to the over-smoothing problem, and eventually harm the model performance. The experimental results also demonstrate this point. From these experiments, an optimal baseline model is obtained, which will be utilized to construct the multi-stream RA-GCN.

For another hyper-parameter $\delta$ mentioned in Section \ref{sssec:activation}, we select its value on a two-stream RA-GCN, with the hyper-parameters $D=2$ and $L=5$. The $\delta$ decides which joints are activated, and hereby controls the number of activated joints for each stream. As seen in the bottom part of Tab.\ref{tab:parameters}, the model obtains the best accuracy when $\delta$ is set to 0.3 on the CS benchmark.

\subsection{Ablation Studies}
\label{ssec:ablation}

The proposed method consists of several fundamental components, e.g., the data preprocessing module, the activation module and so on. In this section, we will analyze the significance of each component. All these experiments are performed by a two-stream RA-GCN with $D=2$, $L=5$ and $\delta=0.3$ on the CS benchmark, and the results are presented in the top line of Tab.\ref{tab:ablation}. As we remove the activation module, the accuracy of our model will drop by 1.2\%. And the pretrained procedure is also important, without which the performance will have a 1.5\% decline. In addition, if the Heaviside step function $\varepsilon(\cdot)$ and the threshold $\delta$ are replaced by a Softmax function as shown in our previous model \cite{song2019richly}, then the accuracy will drop to 85.8\%. According to these experimental results, the necessity of each component in our model is validated for boosting the performance of action recognition.

Furthermore, as seen in the middle line, the data preprocessing module brings a huge improvement, without which the performance is significantly deteriorated. Concretely, the temporal displacements of raw skeletons obtain the best performance, but which are still worse than the whole feature concatenated by the original coordinates, relative coordinates, and temporal displacements. That means all components of the data preprocessing module are beneficial to our model.

The bottom line of Tab.\ref{tab:ablation} shows the results of the RA-GCN models with different numbers of streams. We will find that when the stream number is more than 3, the accuracy growth will be moderate, which is analyzed in Section \ref{ssec:analysis}.

\begin{table}[t]
\caption{Comparison of the SOTA methods on the two benchmarks of NTU 60 in accuracy (\%) and model size (million)}
\label{tab:ntu60}
\centering
\begin{tabular}{cc|cc|cc}
\hline
Model & Year & Param. & Data & CS & CV \\
\hline
DSSCA-SSLM \cite{shahroudy2017deep} & 2017 & -- & Both & 74.9 & -- \\
2D-3D-Softargma \cite{luvizon20182d} & 2018 & -- & RGB & 85.5 & -- \\
Glimpse Clouds \cite{baradel2018glimpse} & 2018 & -- & RGB & 86.6 & 93.2 \\
\hline
\hline
H-BRNN \cite{du2015hierarchical} & 2015 & -- & Skeleton & 59.1 & 64.0 \\
VA-LSTM \cite{zhang2017view} & 2017 & -- & Skeleton & 79.4 & 87.6 \\
CNN+Motion+Trans \cite{li2017skeleton} & 2017 & -- & Skeleton & 83.2 & 89.3 \\
3scale ResNet152 \cite{li2017skeleton2} & 2017 & -- & Skeleton & 85.0 & 92.3 \\
HCN \cite{li2018co} & 2018 & -- & Skeleton & 86.5 & 91.1 \\
\hline
ST-GCN \cite{yan2018spatial} & 2018 & 3.10$^\star$ & Skeleton & 81.5 & 88.3 \\
DPRL+GCNN \cite{tang2018deep} & 2018 & -- & Skeleton & 83.5 & 89.8 \\
SR-TSL \cite{si2018skeleton} & 2018 & 19.07$^\star$ & Skeleton & 84.8 & 92.4 \\
PB-GCN \cite{thakkar2018part} & 2018 & -- & Skeleton & 87.5 & 93.2 \\
AS-GCN \cite{li2019actional} & 2019 & 6.99$^\star$ & Skeleton & 86.8 & 94.2 \\
2s-AGCN \cite{shi2019two} & 2019 & 6.94$^\star$ & Skeleton & 88.5 & 95.1 \\
AGC-LSTM \cite{si2019attention} & 2019 & 22.89$^\dagger$ & Skeleton & {\bf 89.2} & {\bf 95.0} \\
PL-GCN \cite{huang2020part} & 2020 & 20.70$^\dagger$ & Skeleton & 89.2 & 95.0 \\
NAS-GCN \cite{peng2020learning} & 2020 & 6.57$^\dagger$ & Skeleton & 89.4 & 95.7 \\
\hline
preliminary version \cite{song2019richly} & 2019 & 6.21 & Skeleton & 85.9 & 93.5 \\
baseline (1s RA-GCN) & 2020 & 2.03 & Skeleton & 85.8 & 93.1 \\
2s RA-GCN & 2020 & 4.13 & Skeleton & 86.7 & 93.4 \\
3s RA-GCN & 2020 & 6.21 & Skeleton & {\bf 87.3} & {\bf 93.6} \\
\hline
\multicolumn{6}{l}{$^\star$: These results are implemented by ourselves.} \\
\multicolumn{6}{l}{$^\dagger$: These results are provided by their authors.} \\
\end{tabular}
\vspace{-0.4cm}
\end{table}

\begin{table}[t]
\caption{Comparison of the SOTA methods on the two benchmarks of NTU 120 in accuracy (\%) and model size (million)}
\label{tab:ntu120}
\centering
\begin{tabular}{cc|cc|cc}
\hline
Model & Year & Param. & Data & CSub & CSet \\
\hline
Pose Evolution Map \cite{liu2018recognizing} & 2017 & -- & Both & 64.6 & 66.9 \\
Soft RNN \cite{hu2018early} & 2018 & -- & RGB & 36.3 & 44.9 \\
\hline
\hline
PA-LSTM \cite{shahroudy2016ntu} & 2016 & -- & Skeleton & 25.5 & 26.3 \\
ST-LSTM \cite{liu2016spatio} & 2016 & -- & Skeleton & 55.7 & 57.9 \\
2s attention LSTM \cite{liu2017skeleton} & 2017 & -- & Skeleton & 61.2 & 63.3 \\
Skeleton Visualization \cite{liu2018enhanced} & 2018 & -- & Skeleton & 60.3 & 63.2 \\
FSNet \cite{liu2018skeleton} & 2018 & -- & Skeleton & 59.9 & 62.4 \\
SkeleMotion \cite{caetano2019skelemotion} & 2019 & -- & Skeleton & 67.7 & 66.9 \\
TSRJI \cite{caetano2019skeleton} & 2019 & -- & Skeleton & 67.9 & 62.8 \\
\hline
ST-GCN \cite{yan2018spatial} & 2018 & 3.10$^\star$ & Skeleton & 70.7$^\star$ & 73.2$^\star$ \\
SR-TSL \cite{si2018skeleton} & 2018 & 19.07$^\star$ & Skeleton & 74.1$^\star$ & 79.9$^\star$ \\
2s-AGCN \cite{shi2019two} & 2019 & 6.94$^\star$ & Skeleton & {\bf 82.5}$^\star$ & {\bf 84.2}$^\star$ \\
AS-GCN \cite{li2019actional} & 2019 & 6.99$^\star$ & Skeleton & 77.7$^\dagger$ & 78.9$^\dagger$ \\
GVFE+DH-TCN \cite{papadopoulos2019vertex} & 2019 & -- & Skeleton & 78.3 & 79.8 \\
\hline
preliminary version \cite{song2019richly} & 2019 & 6.25 & Skeleton & 74.4 & 79.4 \\
baseline (1s RA-GCN) & 2020 & 2.07 & Skeleton & 78.2 & 80.0 \\
2s RA-GCN & 2020 & 4.17 & Skeleton & 81.0 & 82.5 \\
3s RA-GCN & 2020 & 6.25 & Skeleton & {\bf 81.1} & {\bf 82.7} \\
\hline
\multicolumn{6}{l}{$^\star$: These results are implemented by ourselves.} \\
\multicolumn{6}{l}{$^\dagger$: These results are reported in \cite{papadopoulos2019vertex}.} \\
\end{tabular}
\vspace{-0.4cm}
\end{table}

\subsection{Experimental Results on Standard Dataset}
\label{ssec:results}

We compare the performance of RA-GCN against several previous SOTA methods on the NTU RGB+D 60 \& 120 datasets. The hyper-parameters are chosen as the optimal value given in Tab.\ref{tab:parameters}. Tab.\ref{tab:ntu60} and Tab.\ref{tab:ntu120} display the experimental results of one-stream (1s, baseline), two-stream (2s), three-stream (3s) RA-GCN and the other SOTAs methods, e.g., AGC-LSTM \cite{si2019attention} and so on.

\paragraph{NTU RGB+D 60} Since Yan et al. \cite{yan2018spatial} introduce GCN into skeleton action recognition, there have been many graph-based methods proposed recently, including our methods. Compared to them, our method is only 1.9\% less than AGC-LSTM \cite{si2019attention} on the CS benchmark and 1.4\% on the CV benchmark, while the two accuracy differences are 1.9\% and 1.4\% for PL-GCN \cite{huang2020part} or 2.1\% and 2.1\% for NAS-GCN \cite{peng2020learning}. Although the newly public methods, e.g., PL-GCN and NAS-GCN, have better recognition accuracies than ours, their complexities in model size or training procedures are much higher than the proposed RA-GCN. Compared to ST-GCN \cite{yan2018spatial}, which is the first graph-based model for skeleton action recognition, our method outperforms by 5.8\% and 5.3\%, respectively. With respect to the SOTA methods which is not based on GCN, e.g., VA-LSTM \cite{zhang2017view} and HCN \cite{li2018co}, our method achieves a more significant superiority. The 3s RA-GCN outperforms HCN by 0.8\% and 2.5\%, and exceeds VA-LSTM nearly 8\%.

\paragraph{NTU RGB+D 120} On the currently largest indoor action recognition dataset, NTU RGB+D 120, our approach achieves 81.1\% for the CSub benchmark and 82.7\% for the CSet benchmark. Compared to other GCN-based methods, the proposed method boosts the performance significantly. Concretely, 3s RA-GCN outperforms GVFE+DH-TCN \cite{papadopoulos2019vertex} by 2.8\% and 2.9\% on the two benchmarks, while the non-graph models such as TSRJI \cite{caetano2019skeleton} are far behind our model. Futhermore, for more convincing, three popular models, i.e., ST-GCN \cite{yan2018spatial}, SR-TSL \cite{si2018skeleton} and 2s-AGCN \cite{shi2019two}, are implemented by ourselves, according to their released codes. Compared with them, our model only falls behind 2s-AGCN by 1.4\% and 1.5\% on the two benchmarks of NTU 120 dataset.

\paragraph{Model complexity} To compare the model complexity and computational cost, we calculate the number of parameters in terms of released codes, or ask the authors for the complexities of some models. All the results shown in Tab.\ref{tab:ntu60} and Tab.\ref{tab:ntu120}, from which our model contains similar amount of parameters with other SOTA methods. For some LSTM-based methods, such as SR-TSL \cite{si2018skeleton}, our model only contains nearly 1/3 parameters, due to the efficiency of the GCN technique. Totally, there are only 6.21 million parameters in the 3s RA-GCN, and the inference speed of this model is about 18.7 sequences per second per GPU. Note that there are usually more than 50 frames in an action sequence, thus the training and testing speeds are obviously sufficient for real-time processing.

In general, the proposed RA-GCN can achieve comparable performance with the SOTA methods, because the main purpose of RA-GCN is to discover sufficient redundant representation, while most actions can be recognized by only a few informative joints. However, when these informative joints are occluded or disturbed, the performance of traditional methods will deteriorate significantly.

\subsection{Experimental Results on Occlusion Datasets}
\label{ssec:occlusion}

\begin{table}[t]
\caption{Experimental results (\%) with Frame Occlusion on the CS benchmark of NTU 60 (top) and the CSet benchmark of NTU 120 (bottom)}
\label{tab:frame}
\centering
\begin{tabular}{ccccccc}
\hline
Frame & \multicolumn{6}{c}{Number of Occluded Frames} \\
\cline{2-7}
Occlusion & 0 & 10 & 20 & 30 & 40 & 50 \\
\hline
\hline
ST-GCN \cite{yan2018spatial} & 80.7 & 69.3 & 57.0 & 44.5 & 34.5 & 24.0 \\
SR-TSL \cite{si2018skeleton} & 84.8 & 70.9 & 62.6 & 48.8 & 41.3 & 28.8 \\
2s-AGCN \cite{shi2019two} & {\bf 88.5} & 74.8 & 60.8 & 49.7 & 38.2 & 28.0 \\
preliminary version \cite{song2019richly} & 85.9 & 81.9 & 75.0 & {\bf 66.3} & {\bf 54.4} & {\bf 40.6} \\
baseline (1s RA-GCN) & 85.8 & 81.6 & 72.9 & 61.6 & 47.9 & 34.0 \\
2s RA-GCN & 86.7 & 83.0 & {\bf 76.4} & 65.6 & 53.1 & 39.5 \\
3s RA-GCN & 87.3 & {\bf 83.9} & {\bf 76.4} & {\bf 66.3} & 53.2 & 38.5 \\
difference* & 1.5 & 2.3 & 3.5 & 4.7 & 5.3 & 4.5 \\
\hline
\hline
ST-GCN \cite{yan2018spatial} & 73.2 & 60.8 & 48.8 & 38.2 & 27.3 & 17.4 \\
SR-TSL \cite{si2018skeleton} & 79.9 & 67.4 & 58.8 & 50.4 & 44.7 & 37.1 \\
2s-AGCN \cite{shi2019two} & {\bf 84.2} & 73.4 & 56.8 & 44.2 & 33.0 & 23.4 \\
preliminary version \cite{song2019richly} & 79.4 & 76.2 & 69.8 & 60.4 & 47.4 & 33.0 \\
baseline (1s RA-GCN) & 80.0 & 76.1 & 67.9 & 57.1 & 44.0 & 30.3 \\
2s RA-GCN & 82.5 & 79.1 & 72.4 & {\bf 63.3} & {\bf 51.2} & 36.6 \\
3s RA-GCN & 82.7 & {\bf 79.6} & {\bf 72.9} & {\bf 63.3} & {\bf 51.2} & {\bf 36.8} \\
difference* & 2.7 & 3.5 & 5.0 & 6.2 & 7.2 & 6.5 \\
\hline
\multicolumn{7}{l}{*: the difference between 3s RA-GCN and the baseline model}
\end{tabular}
\vspace{-0.4cm}
\end{table}

\begin{table}[t]
\caption{Experimental results (\%) with Part Occlusion on the CS benchmark of NTU 60 (top) and the CSet benchmark of NTU 120 (bottom)}
\label{tab:part}
\centering
\begin{tabular}{ccccccc}
\hline
Part & \multicolumn{6}{c}{Occluded Part} \\
\cline{2-7}
Occlusion & None & 1 & 2 & 3 & 4 & 5 \\
\hline
\hline
ST-GCN \cite{yan2018spatial} & 80.7 & 71.4 & 60.5 & 62.6 & 77.4 & 50.2 \\
SR-TSL \cite{si2018skeleton} & 84.8 & 70.6 & 54.3 & 48.6 & 74.3 & 56.2 \\
2s-AGCN \cite{shi2019two} & {\bf 88.5} & 72.4 & 55.8 & {\bf 82.1} & 74.1 & 71.9 \\
preliminary version \cite{song2019richly} & 85.9 & 73.4 & 60.4 & 73.5 & 81.8 & 70.6 \\
baseline (1s RA-GCN) & 85.8 & 69.9 & 54.0 & 66.8 & 82.4 & 64.9 \\
2s RA-GCN & 86.7 & {\bf 75.9} & {\bf 62.1} & 69.2 & {\bf 83.3} & {\bf 72.8} \\
3s RA-GCN & 87.3 & 74.5 & 59.4 & 74.2 & 83.2 & 72.3 \\
difference* & 1.5 & 4.6 & 5.4 & 7.4 & 0.8 & 7.4 \\
\hline
\hline
ST-GCN \cite{yan2018spatial} & 73.2 & 59.7 & 47.3 & 52.5 & 68.5 & 48.5 \\
SR-TSL \cite{si2018skeleton} & 79.9 & 59.4 & 50.3 & 41.2 & 64.8 & 55.0 \\
2s-AGCN \cite{shi2019two} & {\bf 84.2} & 62.8 & 46.6 & {\bf 77.8} & 67.0 & 60.7 \\
preliminary version \cite{song2019richly} & 79.4 & 65.6 & 51.2 & 57.3 & 75.3 & 64.9 \\
baseline (1s RA-GCN) & 80.0 & 64.0 & 49.7 & 50.0 & 74.7 & 60.2 \\
2s RA-GCN & 82.5 & 67.4 & 54.1 & 56.0 & 77.6 & 67.7 \\
3s RA-GCN & 82.7 & {\bf 68.5} & {\bf 54.9} & 57.5 & {\bf 79.0} & {\bf 69.9} \\
difference* & 2.7 & 4.5 & 5.2 & 7.5 & 4.3 & 9.7 \\
\hline
\multicolumn{7}{l}{*: the difference between 3s RA-GCN and the baseline model}
\end{tabular}
\vspace{-0.4cm}
\end{table}

\begin{table}[t]
\caption{Experimental results (\%) with Block Occlusion on the CS benchmark of NTU 60 (top) and the CSet benchmark of NTU 120 (bottom)}
\label{tab:block}
\centering
\begin{tabular}{ccccccc}
\hline
Block & \multicolumn{6}{c}{Height Range of the Horizontal Line} \\
\cline{2-7}
Occlusion & None & 1 & 2 & 3 & 4 & 5 \\
\hline
\hline
ST-GCN \cite{yan2018spatial} & 80.7 & 76.4 & 70.1 & 60.7 & 48.4 & 36.1 \\
SR-TSL \cite{si2018skeleton} & 84.8 & 74.9 & 69.3 & 61.1 & 49.4 & 36.9 \\
2s-AGCN \cite{shi2019two} & {\bf 88.5} & 79.2 & 73.6 & 64.6 & 53.4 & 40.2 \\
preliminary version \cite{song2019richly} & 85.9 & 81.6 & 78.6 & 72.5 & 62.3 & 48.2 \\
baseline (1s RA-GCN) & 85.8 & 82.6 & 78.0 & 69.5 & 57.8 & 43.8 \\
2s RA-GCN & 86.7 & 84.4 & {\bf 81.2} & {\bf 74.0} & {\bf 62.9} & {\bf 48.7} \\
3s RA-GCN & 87.3 & {\bf 84.5} & 81.0 & 73.8 & 62.3 & 47.6 \\
difference* & 1.5 & 1.8 & 3.0 & 4.3 & 4.5 & 3.8 \\
\hline
\hline
ST-GCN \cite{yan2018spatial} & 73.2 & 65.5 & 56.8 & 45.0 & 32.7 & 22.6 \\
SR-TSL \cite{si2018skeleton} & 79.9 & 68.2 & 60.1 & 47.7 & 36.5 & 29.0 \\
2s-AGCN \cite{shi2019two} & {\bf 84.2} & 71.7 & 65.7 & 56.3 & 44.6 & 31.1 \\
preliminary version \cite{song2019richly} & 79.4 & 75.7 & 72.3 & 63.8 & 50.9 & 36.0 \\
baseline (1s RA-GCN) & 80.0 & 74.8 & 70.0 & 61.3 & 48.7 & 34.7 \\
2s RA-GCN & 82.5 & 79.0 & 75.2 & 67.8 & 55.8 & 40.6 \\
3s RA-GCN & 82.7 & {\bf 79.7} & {\bf 76.2} & {\bf 69.1} & {\bf 57.3} & {\bf 42.2} \\
difference* & 2.7 & 4.9 & 6.2 & 7.8 & 8.6 & 7.5 \\
\hline
\multicolumn{7}{l}{*: the difference between 3s RA-GCN and the baseline model}
\end{tabular}
\vspace{-0.4cm}
\end{table}

\begin{table}[t]
\caption{Experimental results (\%) with Random Occlusion on the CS benchmark of NTU 60 (top) and the CSet benchmark of NTU 120 (bottom)}
\label{tab:random}
\centering
\begin{tabular}{ccccccc}
\hline
Random & \multicolumn{6}{c}{Occluded Probability} \\
\cline{2-7}
Occlusion & 0 & 0.2 & 0.3 & 0.4 & 0.5 & 0.6 \\
\hline
\hline
ST-GCN \cite{yan2018spatial} & 80.7 & 12.4 & 6.6 & 6.2 & 4.0 & 4.2 \\
SR-TSL \cite{si2018skeleton} & 84.8 & 43.0 & 25.2 & 12.1 & 6.0 & 3.7 \\
2s-AGCN \cite{shi2019two} & {\bf 88.5} & 38.5 & 22.8 & 13.4 & 8.5 & 6.1 \\
preliminary version \cite{song2019richly} & 85.9 & 84.1 & 81.7 & 77.2 & 70.0 & 57.4 \\
baseline (1s RA-GCN) & 85.8 & 82.4 & 77.1 & 72.3 & 63.8 & 49.9 \\
2s RA-GCN & 86.7 & 85.2 & {\bf 83.1} & {\bf 79.4} & {\bf 73.0} & 60.1 \\
3s RA-GCN & 87.3 & {\bf 85.4} & 82.9 & 78.9 & 71.9 & {\bf 61.1} \\
difference* & 1.5 & 3.0 & 5.8 & 6.6 & 8.1 & 11.2 \\
\hline
\hline
ST-GCN \cite{yan2018spatial} & 73.2 & 4.1 & 2.2 & 1.9 & 1.6 & 1.3 \\
SR-TSL \cite{si2018skeleton} & 79.9 & 44.4 & 27.1 & 10.5 & 8.8 & 5.1 \\
2s-AGCN \cite{shi2019two} & {\bf 84.2} & 20.1 & 9.4 & 6.3 & 4.6 & 3.7 \\
preliminary version \cite{song2019richly} & 79.4 & 77.0 & 73.9 & 70.4 & {\bf 62.6} & 42.1 \\
baseline (1s RA-GCN) & 80.0 & 75.1 & 68.4 & 57.4 & 44.7 & 27.6 \\
2s RA-GCN & 82.5 & 79.7 & {\bf 76.2} & {\bf 71.0} & 62.0 & {\bf 48.7} \\
3s RA-GCN & 82.7 & {\bf 79.8} & 75.6 & 68.9 & 58.1 & 43.7 \\
difference* & 2.7 & 4.7 & 7.2 & 11.5 & 13.4 & 16.1 \\
\hline
\multicolumn{7}{l}{*: the difference between 3s RA-GCN and the baseline model}
\end{tabular}
\vspace{-0.4cm}
\end{table}

In this section, we will analyze the experimental results on the synthetic occlusion datasets based on the CS benchmark of NTU 60 dataset and the CSet benchmark of NTU 120 dataset. There are four types of occlusion given as follows:

\paragraph{Frame occlusions} This type of occlusion is constructed to simulate temporal occlusion. We randomly occlude a subsequence in first 100 frames, because the length of most samples is less than 100. The length of the subsequence is set to 10, 20, 30, 40, 50, respectively, and the experimental results are shown in Tab.\ref{tab:frame}. It is observed that the proposed RA-GCN achieves a significant superiority to ST-GCN \cite{yan2018spatial}, SR-TSL \cite{si2018skeleton} and 2s-AGCN \cite{shi2019two}. Besides, the difference between 3s RA-GCN and the baseline model shows a rising trend with the increasing number of occluded frames. As to the comparison of different numbers of streams, 3s RA-GCN is a little better than the others.

\paragraph{Part occlusions} The part occlusion aims at imitating the cases that some key parts of a person are occluded. The occluded parts 1, 2, 3, 4, 5 denote left arm, right arm, two hands, two legs and torso, respectively. As seen in Tab.\ref{tab:part}, there is a huge gap between RA-GCN and other SOTA methods. Compared to the baseline model, 3s RA-GCN obtains large advantages when evaluating without parts 3 and 5. Moreover, 2s RA-GCN has a similar performance with the 3s model.

\paragraph{Block occlusions} Usually, the pedestrian would be occluded by contextual objects. For simulating this real scenario, we design the block occlusion experiments, which occlude the joints behind a predefined horizontal line. The height of the horizontal line is set to five ranges, and the fourth line of Fig.\ref{fig:dataset} shows an example of range 3. Tab.\ref{tab:block} presents these experimental results. Similar with the above occlusion experimental results, our RA-GCN obtains the best accuracies over most of experiments compared to other models.

\paragraph{Random occlusions} During the process of signal transmission, the transmitted data are prone to be lost. Thus, the purpose of random occlusion experiments is to imitate this situation. The occluded probability for every joint is set to 0.2, 0.3, 0.4, 0.5 and 0.6, respectively. The experimental results are displayed in Tab.\ref{tab:random}, from which we can see that RA-GCN extremely alleviates the performance deterioration, while ST-GCN \cite{yan2018spatial}, SR-TSL \cite{si2018skeleton} and 2s-AGCN \cite{shi2019two} have a rapid performance degradation. This phenomenon is caused mainly because these conventional graph-based models require an integrated graph structure, however, which is entirely destroyed by random occlusions.

\paragraph{Failure cases} Specially, we also find that when some important joints, such as right arm, are occluded, some action categories, e.g., {\it handshaking}, cannot be inferred by other joints. The proposed method will fail in such cases.

\subsection{Experimental Results on Jittering Datasets}
\label{ssec:jittering}

\begin{table}[t]
\caption{Experimental results (\%) with jittering skeletons ($\mu=0$, $\sigma=0.1$) on the CS benchmark of NTU 60 (top) and the CSet benchmark of NTU 120 (bottom)}
\label{tab:jittering1}
\centering
\begin{tabular}{ccccccc}
\hline
$\mu=0$ & \multicolumn{6}{c}{Jittering Probability} \\
\cline{2-7}
$\sigma=0.1$ & 0 & 0.02 & 0.04 & 0.06 & 0.08 & 0.10 \\
\hline
\hline
ST-GCN \cite{yan2018spatial} & 80.7 & 66.4 & 44.1 & 32.7 & 13.3 & 7.0 \\
SR-TSL \cite{si2018skeleton} & 84.8 & 70.4 & 53.2 & 41.0 & 33.9 & 21.4 \\
2s-AGCN \cite{shi2019two} & {\bf 88.5} & 74.9 & 60.9 & 41.9 & 29.4 & 20.6 \\
preliminary version \cite{song2019richly} & 85.9 & 73.2 & 59.8 & 45.3 & 41.6 & 34.5 \\
baseline (1s RA-GCN) & 85.8 & 84.1 & 66.1 & 34.2 & 22.2 & 13.9 \\
2s RA-GCN & 86.7 & 70.0 & 55.3 & 48.2 & 41.5 & {\bf 36.4} \\
3s RA-GCN & 87.3 & {\bf 84.2} & {\bf 72.4} & {\bf 61.6} & {\bf 42.4} & 28.7 \\
difference* & 1.5 & 0.1 & 6.3 & 27.4 & 20.2 & 14.8 \\
\hline
\hline
ST-GCN \cite{yan2018spatial} & 73.2 & 63.4 & 50.2 & 33.7 & 18.6 & 10.3 \\
SR-TSL \cite{si2018skeleton} & 79.9 & 60.3 & 50.9 & 39.2 & 30.7 & 19.6 \\
2s-AGCN \cite{shi2019two} & {\bf 84.2} & 42.3 & 37.9 & 35.8 & 31.0 & {\bf 23.7} \\
preliminary version \cite{song2019richly} & 79.4 & 78.1 & {\bf 73.4} & {\bf 55.2} & 26.9 & 17.2 \\
baseline (1s RA-GCN) & 80.0 & 72.2 & 35.9 & 12.6 & 7.0 & 5.4 \\
2s RA-GCN & 82.5 & {\bf 79.3} & 62.5 & 38.4 & 22.2 & 15.2 \\
3s RA-GCN & 82.7 & 77.8 & 65.7 & 47.9 & {\bf 29.5} & 20.5 \\
difference* & 2.7 & 5.6 & 29.8 & 35.3 & 22.5 & 15.1 \\
\hline
\multicolumn{7}{l}{*: the difference between 3s RA-GCN and the baseline model}
\end{tabular}
\vspace{-0.4cm}
\end{table}

\begin{table}[t]
\caption{Experimental results (\%) with jittering skeletons ($\mu=0$, $\sigma=0.05$) on the CS benchmark of NTU 60 (top) and the CSet benchmark of NTU 120 (bottom)}
\label{tab:jittering2}
\centering
\begin{tabular}{ccccccc}
\hline
$\mu=0$ & \multicolumn{6}{c}{Jittering Probability} \\
\cline{2-7}
$\sigma=0.05$ & 0 & 0.02 & 0.04 & 0.06 & 0.08 & 0.10 \\
\hline
\hline
ST-GCN \cite{yan2018spatial} & 80.7 & 76.4 & 65.1 & 50.2 & 32.8 & 19.5 \\
SR-TSL \cite{si2018skeleton} & 84.8 & 69.4 & 55.3 & 50.1 & 46.6 & 39.2 \\
2s-AGCN \cite{shi2019two} & {\bf 88.5} & 78.9 & 79.8 & 76.8 & 72.6 & 60.7 \\
preliminary version \cite{song2019richly} & 85.9 & 83.8 & 81.3 & 75.3 & 69.2 & {\bf 61.4} \\
baseline (1s RA-GCN) & 85.8 & 82.4 & 77.1 & 72.3 & 63.8 & 49.9 \\
2s RA-GCN & 86.7 & 83.8 & 77.3 & 71.6 & 61.6 & 58.5 \\
3s RA-GCN & 87.3 & {\bf 87.0} & {\bf 84.5} & {\bf 81.1} & {\bf 72.9} & {\bf 61.4} \\
difference* & 1.5 & 4.6 & 7.4 & 8.8 & 9.1 & 11.5 \\
\hline
\hline
ST-GCN \cite{yan2018spatial} & 73.2 & 70.4 & 64.0 & 59.5 & 44.0 & 32.1 \\
SR-TSL \cite{si2018skeleton} & 79.9 & 68.2 & 55.4 & 47.9 & 41.3 & 33.6 \\
2s-AGCN \cite{shi2019two} & {\bf 84.2} & 56.7 & 49.2 & 41.9 & 37.3 & 32.0 \\
preliminary version \cite{song2019richly} & 79.4 & 79.0 & 78.7 & {\bf 77.1} & {\bf 72.3} & {\bf 61.2} \\
baseline (1s RA-GCN) & 80.0 & 79.4 & 71.9 & 46.4 & 25.2 & 14.3 \\
2s RA-GCN & 82.5 & 81.9 & {\bf 79.7} & 69.5 & 44.9 & 31.8 \\
3s RA-GCN & 82.7 & {\bf 82.0} & 79.6 & 74.2 & 62.1 & 48.5 \\
difference* & 2.7 & 2.6 & 7.7 & 27.8 & 36.9 & 34.2 \\
\hline
\multicolumn{7}{l}{*: the difference between 3s RA-GCN and the baseline model}
\end{tabular}
\vspace{-0.4cm}
\end{table}

To discuss the impact of jittering skeletons, two jittering datasets are designed by adding different Gaussian noises, and the corresponding experimental results are shown in Tab.\ref{tab:jittering1} and Tab.\ref{tab:jittering2}. The jittering probability for every joint is set to 0.02, 0.04, 0.06, 0.08 and 0.10, respectively. It is clearly observed from these tables that, the 3s RA-GCN outperforms other models by a huge gap on NTU 60-based jittering dataset. Moreover, with the increase of jittering probability, the gap between 3s RA-GCN and the baseline is synchronously increasing. Therefore, it is proven that our method is robustness to the current skeletons.

\begin{figure}[t]
    \centering
    \includegraphics[width=8.5cm]{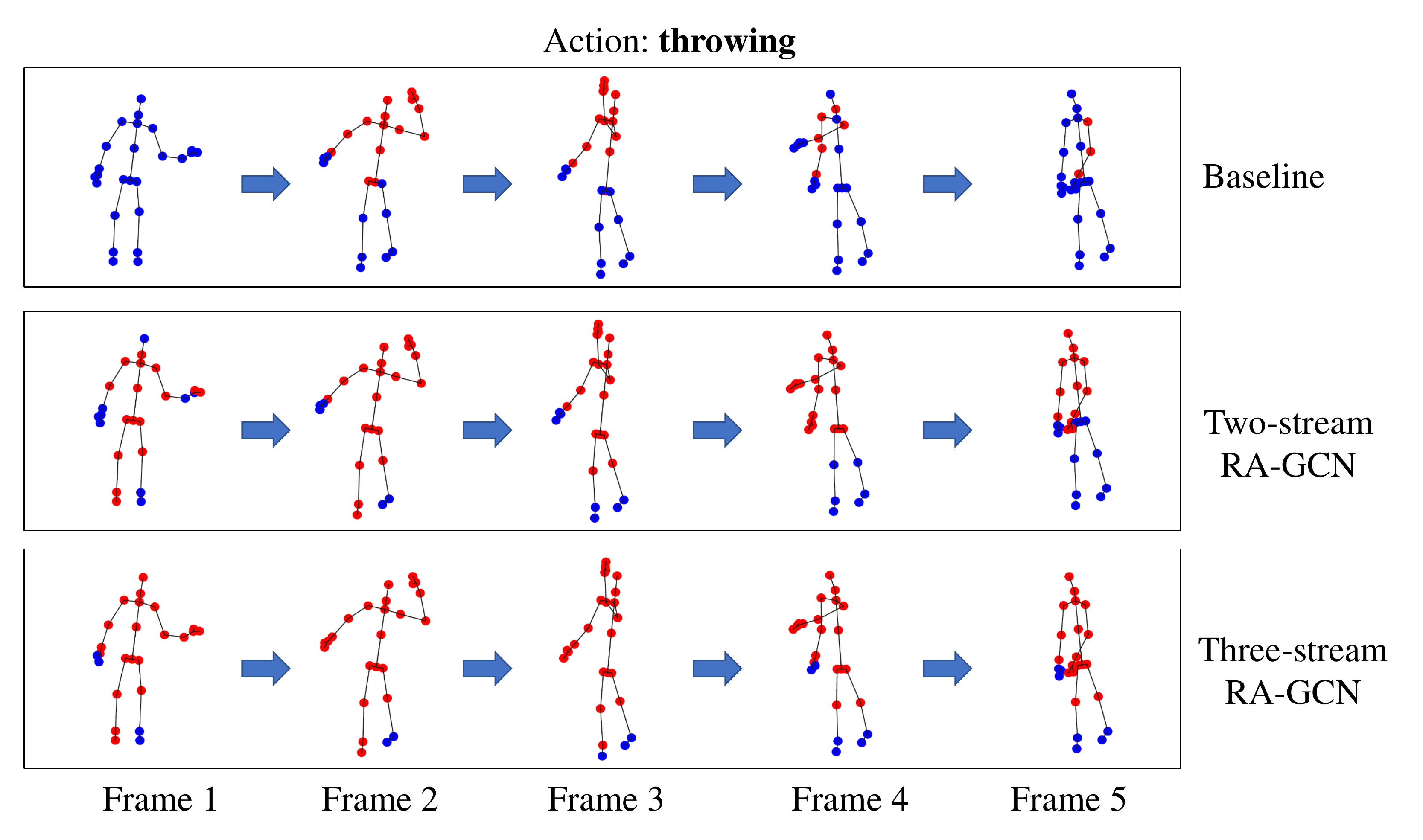}
    \caption{(Best viewed in color.) An example of activated joints of all streams for the baseline model, the RA-GCN with two streams and three streams. The {\color{red}red} points denote the activated joints, while the {\color{blue}blue} points denote the unactivated joints.}
    \label{fig:activated}
\end{figure}

\subsection{Results Analysis}
\label{ssec:analysis}

In this section, we will analyze why our model is more robust to noisy or incomplete input data. Fig.\ref{fig:activated} shows an example of the activated joints. In this figure, the top line shows the activated joints of the baseline model, which is also considered as a 1s RA-GCN. The activated joints in the middle line correspond to a 2s RA-GCN, while the bottom line presents the results of a 3s RA-GCN. For a clearer display, we select five contextual frames from the sequence to represent the action {\it throwing}, instead of a whole sequence.

From this figure, it is observed that the proposed RA-GCN successfully expands the activation map of the baseline model. The baseline model only captures the most discriminative joints, e.g., the joints in two arms. In contrast, our model not only concentrates on the joints in two arms, but also activates some other discriminative joints, which play an auxiliary role in this action, such as the slightly swaying body and stepping legs. Moreover, with the increase of streams, more activated joints can be discovered accordingly. Therefore, the multi-stream RA-GCN still has a modest recognition performance when a few joints are occluded or disturbed. This will lead to a more robust capability of our model to data degradation than other models.

In addition, it is worth to notice that more streams in the RA-GCN will not always obtain a more accurate model, because the number of discriminative joints in an action category is often limited. In this work, three streams are sufficient to discover these joints, and the experimental results in Section \ref{ssec:ablation} also demonstrates this point.

\section{Conclusion}
\label{sec:conclusion}

In this paper, to reduce the impact of noisy or incomplete skeletons in action recognition, we have proposed a novel model named RA-GCN for discovering rich features over all skeleton joints, which achieves a much better performance than the baseline model and improves the robustness of the model. With extensive experiments on the NTU RGB+D 60 \& 120 datasets, we verify the effectiveness and robustness of our model. For evaluating the model's performance on non-standard skeletons, we construct various synthetic datasets composed of four types of occlusion and two types of jittering. On these synthetic datasets, the proposed RA-GCN outperforms the other SOTA methods, as well as showing a significant improvement than the baseline model. In the future, we will consider to add the attention module within our model, in order to make each stream focus more on informative joints.

{\small
    \bibliographystyle{IEEEtran}
    \bibliography{conferences_abrv,IEEE}
}

\begin{IEEEbiography}[{\includegraphics[width=1in,height=1.25in,clip,keepaspectratio]{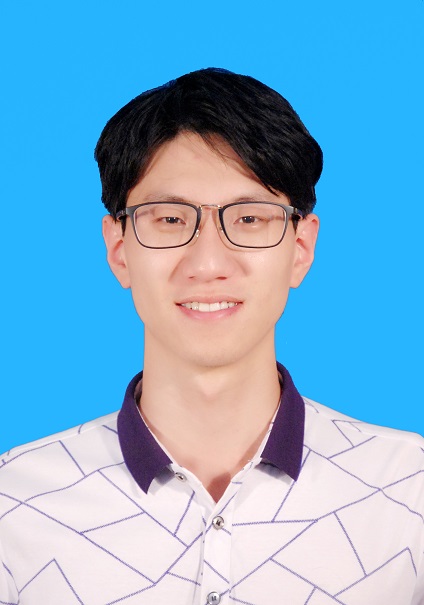}}]{Yi-Fan Song}
received his M.S. degree in Zhengzhou University, Zhengzhou, China, in 2018. Currently, He is a student of the School of Artificial Intelligence, University of Chinese Academy and Sciences (UCAS). And he is working toward the Ph.D. degree in the Center for Research on Intelligent Perception and Computing (CRIPAC), Institute of Automation, Chinese Academy of Sciences (CASIA), Beijing, China. His research interests include computer vision, activity recognition, video surveillance, and time series analysis.
\end{IEEEbiography}

\begin{IEEEbiography}[{\includegraphics[width=1in,height=1.25in,clip,keepaspectratio]{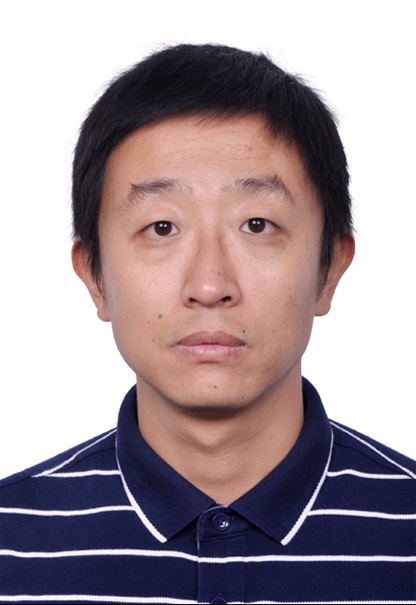}}]{Zhang Zhang}
received the B.S. degree in computer science and technology from Hebei University of Technology, Tianjin, China, in 2002, and the Ph.D. degree in pattern recognition and intelligent systems from the National Laboratory of Pattern Recognition, Institute of Automation, Chinese Academy of Sciences, Beijing, China in 2009. Currently, he is an associate professor at the National Laboratory of Pattern Recognition, Institute of Automation, Chinese Academy of Sciences (CASIA). His research interests include activity recognition, video surveillance, and time series analysis. He has published 20s research papers on computer vision and pattern recognition, including IEEE TPAMI, CVPR, and ECCV etc.
\end{IEEEbiography}

\begin{IEEEbiography}[{\includegraphics[width=1in,height=1.25in,clip,keepaspectratio]{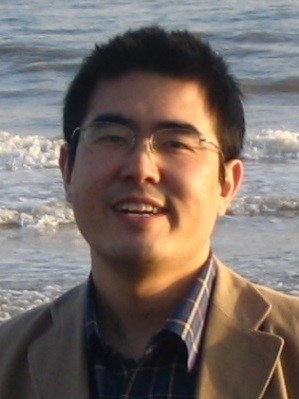}}]{Caifeng Shan}
received the B.Eng. degree from the University of Science and Technology of China (USTC), the M.Eng degree from the Institute of Automation, Chinese Academy of Sciences, and the PhD degree in computer vision from Queen Mary, University of London. His research interests include computer vision, pattern recognition, image and video analysis, machine learning, bio-medical imaging, and related applications. He has authored more than 100 papers and 60 patent applications. He has served as Associate Editor or Guest Editor for many scientific journals including IEEE Transactions on Circuits and Systems for Video Technology and IEEE Journal of Biomedical and Health Informatics. He is Senior Member of IEEE.
\end{IEEEbiography}

\begin{IEEEbiography}[{\includegraphics[width=1in,height=1.25in,clip,keepaspectratio]{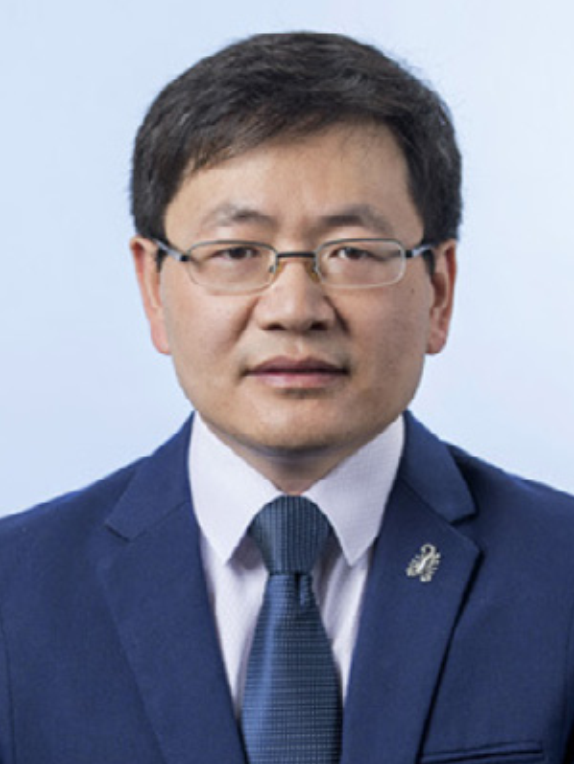}}]{Liang Wang}
received both the BEng and MEng degrees from Anhui University in 1997 and 2000, respectively, and the PhD degree from the Institute of Automation, Chinese Academy of Sciences (CASIA) in 2004. From 2004 to 2010,he was a research assistant at Imperial College London, United Kingdom, and Monash University, Australia, a research fellow at the University of Melbourne, Australia, and a lecturer at the University of Bath, United Kingdom, respectively. Currently, he is a full professor of the Hundred Talents Program at the National Lab of Pattern Recognition, CASIA. His major research interests include machine learning, pattern recognition, and computer vision. He has widely published in highly ranked international journals such as IEEE Transactions on Pattern Analysis and Machine Intelligence and IEEE Transactions on Image Processing, and leading international conferences such as CVPR, ICCV, and ICDM. He is an IEEE Fellow, and an IAPR Fellow.
\end{IEEEbiography}

\end{document}